%% file: main.tex
\documentclass[10pt,twocolumn,letterpaper]{article}

\usepackage{cvpr}

\input{preamble}

\definecolor{cvprblue}{rgb}{0.21,0.49,0.74}
\usepackage[pagebackref,breaklinks,colorlinks,allcolors=cvprblue]{hyperref}

\def\confName{CVPR}
\def\confYear{2027}

\title{Slide2PPT: Native PowerPoint Reconstruction with Localization-Guided Refinement}

\author{
    Baoliang Chen$^{1}$ \quad
    Xinlong Bu$^{2}$ \quad
    Lingyu Zhu$^{1}$ \quad
    Xiangjie Sui$^{3}$ \\[0.3em]
    {\small
    $^{1}$City University of Hong Kong \quad
    $^{2}$South China Normal University \quad
    $^{3}$City University of Macau
    }
}

\makeatletter
\apptocmd{\@maketitle}{\input{sec/teaser}}{}{\PackageError{Slide2PPT}{Could not attach the teaser to the title block}{}}
\AtBeginDocument{%
  \raggedbottom
  \AddToHookNext{shipout/after}{%
    \global\let\@textbottom\relax
    \global\let\@texttop\relax
  }%
}
\makeatother

\begin{document}
\maketitle
\input{sec/00_abstract}

\input{sec/01_introduction}
\input{sec/02_related_work}
\input{sec/03_benchmark}
\input{sec/04_method}
\input{sec/05_experiments}
\input{sec/07_discussion}
\input{sec/06_conclusion}

\FloatBarrier
{
    \small
    \bibliographystyle{ieeenat_fullname}
    \bibliography{main}
}


\end{document}

%% file: preamble.tex
\usepackage{multirow}
\usepackage{colortbl}
\definecolor{PaperHeader}{HTML}{F7E8B7}
\definecolor{PaperFocus}{HTML}{EEEAF7}
\definecolor{PaperBest}{HTML}{B3262E}
\definecolor{PaperSecond}{HTML}{245B91}
\usepackage{tabularx}
\usepackage{tikz}
\newlength{\metriclabelwidth}
\newlength{\metricgroupoffset}
\definecolor{PaperDivider}{gray}{0.55}

\definecolor{QwenPending}{HTML}{E00000}

\usepackage{placeins}

%% file: sec/teaser.tex
\vspace{-15pt} 
\begin{minipage}{\textwidth}
  \captionsetup{hypcap=false}
  \hypersetup{citecolor=black}
  \centering
  \includegraphics[width=\textwidth]{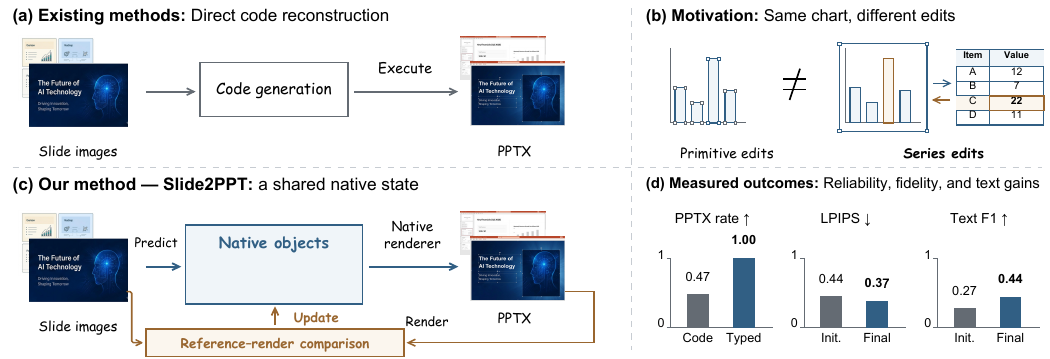}
  \captionof{figure}{Native objects connect reconstruction to correction. (a) Code prediction generates content and drawing logic together. (b) Similar-looking charts can expose different editing interfaces. (c) Comparing reference and rendered images guides correction of the shared native state. (d) Readable-PPTX rate on $108$ inputs across ten configurations, and mean LPIPS/text-box F1 on $80$ scientific figures across five API configurations.}
  \label{fig:graphical_abstract}
\end{minipage}
\par\vspace{12pt}

%% file: sec/00_abstract.tex

\begin{abstract}
Image-to-slide reconstruction must recover both visual appearance and native object structure for subsequent editing. Yet visually faithful charts and tables may lack editable series and cells, and direct code generation couples content prediction to rendering implementation. We present Slide2PPT, a model-agnostic, training-free framework that equips both open- and closed-source vision-language models with native PowerPoint reconstruction capabilities. Through a unified typed interface, the model predicts object records while a fixed renderer materializes their native properties, enabling integration with different backbones without task-specific training or architectural changes. Localization-guided refinement then connects visual discrepancies to object records for correction. Regional updates preserve unselected records, while whole-slide updates address cross-region relationships. Record refinement and targeted object editing support structural and attribute-level changes, with rendered feedback governing iterative acceptance. To evaluate this framework across diverse reconstruction demands, we introduce Slide2PPT-Bench, covering controlled elements, real slides, and scientific figures. Evaluation separately measures file production, rendered agreement, and native-object availability. Experiments across open- and closed-source backbones demonstrate substantially more reliable PPTX generation than direct code generation, improved layout and text recovery on scientific figures, and native chart and table interfaces beyond geometric primitives.
\end{abstract}

%% file: sec/01_introduction.tex
\section{Introduction}

A slide image preserves what readers see but hides what authors edit. A reconstructed chart may look convincing while offering only movable rectangles, with no series values to change. A table may be legible yet lack cells into which a new row can be inserted. Useful image-to-PowerPoint reconstruction therefore requires both a faithful rendering and object properties that support the intended edits.

Recent methods recover slide content through layout-aware code generation~\cite{tang2025slidecoder} and typed region descriptions~\cite{gonzalez2026images2slides}. Native presentation state and rendered feedback also support deck editing~\cite{zheng2026slideforge}. These developments leave a practical question for fixed-image reconstruction: how can a visible error be translated into a change to the responsible editable objects? In a generated drawing program, content prediction is coupled to rendering implementation. Moreover, visual agreement cannot distinguish a native chart from a similar arrangement of drawing primitives.

We introduce Slide2PPT, a model-agnostic, training-free framework that uses native PowerPoint objects as the shared state for reconstruction and correction (Fig.~\ref{fig:graphical_abstract}). An off-the-shelf vision-language model (VLM) serves as an interchangeable backbone to predict typed records, and a fixed renderer implements their native properties, including table cells and chart series. Correction revises those records through the same renderer. This design separates the prediction of slide content from its drawing implementation and connects visible discrepancies to the properties available for editing.

The spatial scope of a correction matters. Changing all records to repair a small label allows unrelated content to change, while restricting edits to a region can exclude an object or connection crossing its boundary. Slide2PPT first corrects records selected around a visual discrepancy, then revisits the full slide. At both scopes, record refinement permits coupled changes and object editing expresses targeted operations. Each candidate is rendered and assessed before entering the working state.

We evaluate this design on our constructed Slide2PPT-Bench, a collection of $1{,}080$ images covering controlled elements, real slides, and scientific figures. On  samples  across ten configurations, typed reconstruction produces readable PPTX files in all cases, compared with a rate of $0.47$ for direct code. On the most challenging scientific figures across five VLM  configurations, correction raises mean text-box F1 from $0.27$ to $0.44$ and lowers page CER from $0.51$ to $0.43$. Separate measurements of file production, rendered agreement, and native-object inventories distinguish the reliability of reconstruction from the properties available for editing.

%% file: sec/02_related_work.tex
\section{Related Work}
\label{sec:related}

\paragraph{Reconstruction from a visual target.}
Image-based slide reconstruction must choose how to represent the recovered content. SlideCoder~\cite{tang2025slidecoder} uses layout decomposition and retrieval to generate Python-PPTX programs. Typed regions in Images2Slides~\cite{gonzalez2026images2slides} support text and image reconstruction through the Google Slides API, while SliDer~\cite{hazimeh2025slider} combines layout priors with iterative SVG refinement. Slide2PPT represents the slide as native PPTX objects, including tables and charts with editable cells and series. The same records carry both the initial prediction and subsequent corrections.

\paragraph{Structured generation and editing.}
Document-to-slide methods use structure to organize content and support revision. DOC2PPT~\cite{fu2022doc2ppt} generates slides from documents, PPTAgent~\cite{zheng2025pptagent} edits reference presentations, and AutoPresent~\cite{ge2025autopresent} generates structured visuals through tool-augmented programs. Rendered feedback also supports native deck editing in SlideForge~\cite{zheng2026slideforge} and step-level critique and validation in ReDeck~\cite{tian2026redeck}. Our focus is reconstruction against a fixed image: local discrepancies select native records for correction, followed by whole-slide updates that can revise relationships across region boundaries.

\paragraph{Reconstruction evaluation.}
Visual similarity and structural correctness impose different requirements on reconstruction. Design2Code~\cite{si2024design2code} studies screenshot-to-code generation, while ChartMimic~\cite{shi2024chartmimic} evaluates chart reconstruction. The layout-and-design PPTBench~\cite{huang2025pptbench} covers slide understanding, modification, and generation. A separate PPTBench~\cite{wang2026pptbenchreconstruction} tests scientific-diagram reconstruction with native-object and semantic criteria, and ReFigBench~\cite{fan2026refigbench} combines structural audits with human comparison. We measure file production, rendered agreement, and native-object availability separately. These measurements distinguish reconstruction failures from the editing interfaces present in each output.

%% file: sec/03_benchmark.tex
\section{Task and Benchmark}
\label{sec:benchmark}

Slide2PPT-Bench evaluates how much of a slide image can be recovered as an editable document. Its three sources span isolated object properties, presentation layouts, and dense scientific graphics. We distinguish a successful rendering from the native structure available for later edits, since either can be recovered without the other.

\subsection{Task Definition}

Given a slide image $I$, we construct a single-slide PowerPoint document $\mathcal{P}$ that resembles $I$ and exposes object-specific editing properties. Exact source recovery is generally impossible: the image hides object hierarchy, occluded content, and source data. A visible chart admits several reconstructions, from a picture to drawing primitives or a native chart, each exposing different editing operations.

The target therefore extends beyond movable shapes. Text boxes expose content and typography, connectors expose endpoints and attachments, and pictures support cropping and replacement. Native tables and charts additionally expose cells and series data for subsequent edits.

\subsection{Benchmark Construction}
\label{sec:data}

\begin{figure}[t]
  \centering
  \includegraphics[width=\linewidth]{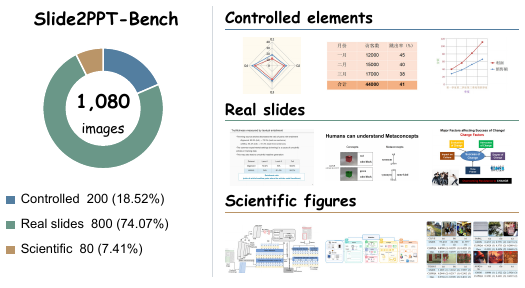}
  \caption{Source proportions and examples in Slide2PPT-Bench. Rows show controlled elements, real slides, and scientific figures. Initialization reconstructions illustrate content diversity; they are not reference images or quality comparisons.}
  \label{fig:benchmark_sources}
\end{figure}

The collection contains $1{,}080$ images from three sources (Fig.~\ref{fig:benchmark_sources}). Each source emphasizes a different reconstruction demand, without defining a calibrated difficulty level. The experiments use different subsets to study full-collection initialization, initialization strategies, and subsequent correction, as specified in Sec.~\ref{sec:settings}.

\paragraph{Controlled elements.}
We generate content attributes with language models and sample categorical properties from Python-PPTX enumerations. JSONL2PPTX renders approximately $10{,}000$ synthetic specifications, from which we randomly select $200$ samples. These samples isolate common object types and attributes, but share the reconstruction renderer and do not test unseen authoring styles.

\paragraph{Real slides.}
We collect $800$ slides from DOC2PPT~\cite{fu2022doc2ppt}, the layout-and-design PPTBench~\cite{huang2025pptbench}, and SlidesBench~\cite{ge2025autopresent}. Their compositions require recovering object placement and stacking alongside content and style, extending evaluation beyond isolated elements.

\paragraph{Scientific figures.}
We collect $80$ scientific-figure screenshots through GPT-assisted retrieval and manual capture. Dense plots and technical diagrams combine small labels with graphical marks and interdependent components. Correction and tool comparisons use these figures.

\subsection{Evaluation Dimensions}
\label{sec:evaluation_dimensions}

We separate file production, visual agreement, and native-object availability. A readable PPTX establishes that the file can be opened. Image, detected-layout, and OCR measurements assess correspondence to the reference. Object inventories record the native interfaces present in that file. These measurements are complementary: a faithful rendering can remain rasterized, while a native chart can contain incorrect values. Editing correctness requires matching source structure and executing edits.

%% file: sec/04_method.tex
\section{Method}
\label{sec:method}

\begin{figure*}[t]
  \centering
  \includegraphics[width=\textwidth]{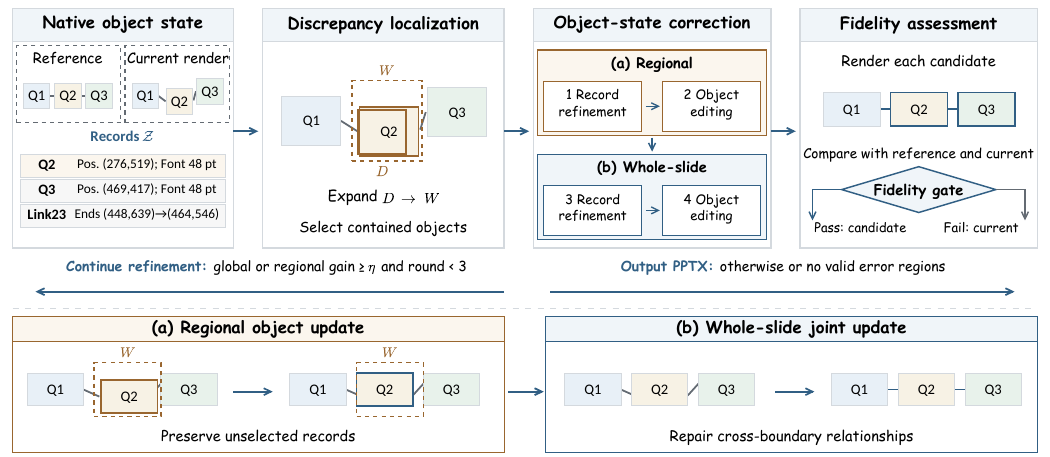}
  \caption{Iterative correction after bootstrap. Discrepancy $D$ expands to window $W$ for record selection. Steps $1$--$4$ apply record refinement and object editing regionally, then across the whole slide. Each candidate is rendered and gated before the next step uses the retained state. Panels (a) and (b) illustrate this progression. Positions use normalized slide coordinates and font sizes are in points.}
  \label{fig:framework}
\end{figure*}

A visual discrepancy identifies where a reconstruction is wrong, but not which document objects should change. Slide2PPT makes this correspondence explicit through a shared native state (Fig.~\ref{fig:framework}). Discrepancies select records for regional correction, preserving unselected records. Whole-slide correction then revisits dependencies that the regional boundary excludes. Both scopes revise the same state and assess the resulting rendering before retaining a change.

\subsection{A Shared Native State}
\label{sec:renderer}
\label{sec:representation}

We represent a slide as an ordered JSONL sequence $\mathcal{Z}=(z_0,\ldots,z_N)$. The background is $z_0$; each foreground record specifies an identifier, geometry, and attributes for a textbox, shape, image, connector, table, or chart. Record order determines back-to-front insertion, and coordinates are normalized per axis to $[0,1000]$. Type templates define fields for cells, series, and connector attachments.

A fixed renderer $\mathcal{G}$ maps these records to native PowerPoint properties. Office rasterization $\mathcal{R}$ then gives $\hat I=\mathcal{R}(\mathcal{G}(\mathcal{Z};I))$. The model predicts slide content and attributes, while the renderer supplies their drawing implementation. Correction can therefore change a table cell or chart series through its record, without generating a new implementation of that object.

Initialization first predicts foreground boxes, then records conditioned on $I$, the boxes, and type templates. Boxes guide spatial prediction without fixing the object decomposition: record prediction can adjust, merge, or split detections. A whole-slide bootstrap applies record refinement followed by object editing (Sec.~\ref{sec:correction_forms}). Iteration then reuses these two forms at regional and whole-slide scopes.

Picture objects retain content without decomposition into primitives. If an image record has no external asset, the renderer crops that content from $I$. Generated backgrounds, when enabled, are also retained during correction, keeping the visual context fixed while foreground records change.

\subsection{Discrepancy-Guided Record Selection}
\label{sec:initialization}
\label{sec:refinement}

Regional correction restricts changes to objects near a visual error. A model compares $I$ with $\hat I$ and predicts discrepancy boxes. Overlapping boxes merge into regions $D$, each expanded to a context window $W$. We select foreground records fully contained in $W$. Let $g(z_i)$ be an object's bounding rectangle, or the rectangle containing a connector's endpoints, and let $\mathcal{J}$ contain all foreground identifiers. Selection and merging satisfy
\begin{equation}
  \begin{aligned}
    &\mathcal{J}_W=\{\operatorname{id}(z_i):i>0,\ g(z_i)\subseteq W\},\\
    &\pi_{\mathcal{J}\setminus\mathcal{J}_W}(\mathcal{Z}')
      =\pi_{\mathcal{J}\setminus\mathcal{J}_W}(\mathcal{Z}).
  \end{aligned}
  \label{eq:regional_update}
\end{equation}
Here $\pi$ extracts ordered records by identifier, and $\mathcal{Z}'$ is a merged candidate. The second line preserves the values and relative order of unselected records. Full containment excludes boundary-crossing objects; a connector is available only when both endpoints lie in $W$. Whole-slide correction later removes this selection restriction.

The selected sequence $\mathcal{Z}_W$ retains full-slide coordinates and is paired with reference and current-render crops. Merging replaces records by identifier, removes omitted selected records, and inserts additions relative to neighboring records. New records and their references are renamed to resolve identifier collisions. The background remains fixed. This constraint preserves unselected records, not every pixel outside $W$: changed geometry, stacking, or attachments can alter the rendered context.

\subsection{Record Refinement and Object Editing}
\label{sec:correction_forms}

A selected region can require a coupled revision or a targeted change. We use two correction forms to express these alternatives. \emph{Record refinement} jointly predicts a revised record sequence, allowing content, attributes, and object composition to change together. \emph{Object editing} predicts ordered operations addressed to individual records, such as changing a property or inserting an object. Both forms operate on native records during bootstrap and iteration.

For a regional update, let $C_W=(I|_W,\hat I|_W,W)$ denote the reference crop, current crop, and window coordinates. Record refinement predicts a sequence through $f_{\rm state}$; object editing predicts operations $(a_1,\ldots,a_L)$. The candidate state is
\begingroup
\small
\setlength{\abovedisplayskip}{4pt}
\setlength{\belowdisplayskip}{4pt}
\setlength{\abovedisplayshortskip}{4pt}
\setlength{\belowdisplayshortskip}{4pt}
\begin{equation}
  \begin{aligned}
    &\widetilde{\mathcal{Z}}_W=
    \begin{cases}
      f_{\rm state}(\mathcal{Z}_W,C_W),&\text{Record refinement},\\
      (\mathcal{A}_{a_L}\circ\cdots\circ\mathcal{A}_{a_1})
      (\mathcal{Z}_W),&\text{Object editing},
    \end{cases}\\
    &\mathcal{Z}'=\mathcal{M}_W(\mathcal{Z},\widetilde{\mathcal{Z}}_W).
  \end{aligned}
  \label{eq:update_forms}
\end{equation}
\endgroup
The operator $\mathcal{A}_{a}$ applies an addressed edit and skips invalid operations. The merge $\mathcal{M}_W$ obeys Eq.~\eqref{eq:regional_update}. For object editing, the complete operation list is executed before rendering and acceptance. Feedback therefore assesses the combined candidate rather than each atomic operation.

Each iteration applies record refinement followed by object editing regionally, then repeats this pair across the whole slide. Regional requests within a stage share a baseline, but proposals are merged and assessed sequentially against the latest accepted state. Whole-slide updates receive the full image pair and all records, allowing them to revise boundary-crossing objects and relationships excluded from regional updates.

\subsection{Scope-Matched Visual Acceptance}
\label{sec:acceptance}

The acceptance test must distinguish a small local improvement from a change affecting the whole composition. A full-image score can dilute a repaired label, while a local score can miss damage elsewhere. After native rendering and Office rasterization, we assess each candidate on image support matched to its update scope. For support $S$, define
\begingroup
\small
\begin{equation}
  e_S(\hat I)=
  \begin{bmatrix}
    1-\operatorname{SSIM}(I|_S,\hat I|_S)\\
    \operatorname{LPIPS}(I|_S,\hat I|_S)\\
    \operatorname{FLIP}(I|_S,\hat I|_S)
  \end{bmatrix},
  \quad d_S=e_S(\hat I)-e_S(\hat I').
  \label{eq:visual_evidence}
\end{equation}
\endgroup
Positive components of $d_S$ indicate improvement. Regional proposals are tested on $D$, excluding the context padding in $W$, and whole-slide proposals on canvas $\Omega$. With minimum gain $\epsilon$ and regional regression tolerance $\tau$, acceptance is
\begin{equation}
  \begin{aligned}
    A_D&=\mathbf{1}\!\left[
      \sum_{j=1}^{3}\mathbf{1}[d_{D,j}\geq\epsilon]\geq2
      \ \land\ \min_j d_{D,j}\geq-\tau\right],\\
    A_\Omega&=\mathbf{1}\!\left[
      \frac{1}{3}\sum_{j=1}^{3}d_{\Omega,j}\geq\frac{\epsilon}{3}\right].
  \end{aligned}
  \label{eq:scope_acceptance}
\end{equation}
Regional acceptance requires improvement in at least two metrics and bounds regression in the third. Whole-slide acceptance uses their mean error reduction. A passing candidate replaces the working state; a rejected candidate leaves it unchanged. These are visual selection rules, without a guarantee of semantic correctness. Thresholds and additional stage-retention rules appear in Supplement Sec.~S2.

The loop re-localizes errors from the retained rendering. It continues only while valid regions remain, at least one stage is accepted, the iteration budget remains, and either the full-slide or mean regional composite gain reaches $\eta=2\times10^{-3}$. It stops after at most three iterations, or earlier if localization fails. A regional stage can be retained despite a declining full-image composite, so individual metrics need not improve monotonically. The exported document is the latest accepted state.

%% file: sec/05_experiments.tex
\section{Experiments}
\label{sec:experiments}

The experiments examine whether the native state produces reliable files, whether rendered feedback improves reconstruction, and what editing interfaces the outputs expose. We first evaluate initial reconstruction across the full collection, then study correction and compare tools on scientific figures. Initialization-strategy and cumulative-stage comparisons help interpret these results.

\subsection{Settings and Measurements}
\label{sec:settings}

Initial reconstruction uses four VLM API and five local configurations (Table~\ref{tab:initialization_summary}). A representative $108$-image subset compares initialization strategies with these configurations plus Qwen3.8-Max~\cite{qwen2026qwen38max}. Correction uses all $80$ challenging scientific figures and the five API configurations; external-tool comparisons use the same figures with GPT-5.6 Sol. Tables report configuration scores or cross-model mean $\pm$ sample SD.

Local inputs are reduced to height $720$ when necessary. API inputs retain their resolution subject to request limits, and API runs enable background generation that local runs omit. WPS Office rasterizes PPTX files at $168$ DPI. These deployment differences are retained in the reported comparisons, without treating API and local scores as a controlled comparison of model capability.

We compare three initialization strategies: \textsc{Code} predicts a Python-PPTX program, \textsc{JSONL} predicts native records, and \textsc{BBox-JSONL} conditions record prediction on detected boxes. BBox-JSONL initializes Slide2PPT. On scientific figures, the cumulative states add whole-slide record refinement and object editing during bootstrap, followed by iterative correction (Sec.~\ref{sec:correction_forms}). 

Appearance is measured by SSIM~\cite{1284395}, LPIPS~\cite{zhang2018perceptual}, FLIP~\cite{andersson2020flip}, CLIP image similarity~\cite{radford2021learning}, and set-level FID~\cite{heusel2017gans}. Layout and text scores compare DocLayout-YOLO detections~\cite{zhao2024doclayout} and PaddleOCR outputs~\cite{cui2025paddleocr30technicalreport} between the reference and reconstruction. Region and text-box F1 use IoU$\geq0.50$, and page CER includes unmatched text and reading-order differences. PPTX rate is the fraction yielding a readable file, in $[0,1]$. Supplement Secs.~S3--S4 define the metrics and failure handling.

\subsection{Full-Dataset Reconstruction}

A fixed renderer supplies the object interface, but reconstruction still depends on the model's visual predictions. Table~\ref{tab:initialization_summary} reports initial BBox-JSONL reconstruction on all $1{,}080$ images. GPT-5.6 Sol leads the API group in appearance, layout, and text agreement. Among local configurations, Qwen3.8-27B gives the strongest results on most metrics, including SSIM $0.78$ and text-box F1 $0.44$. Within-group variation shows that native representation alone does not ensure faithful content or placement. The following experiments test how correction changes this initial state.

\begin{table*}[t]
  \centering
  \caption{Full-dataset reconstruction on $1{,}080$ images: all $14$ quality metrics for each configuration using bbox-guided JSONL initialization. Bold marks within-group optima. API and local deployments differ in resolution and background processing.}
  \label{tab:initialization_summary}
  \small
  \setlength{\tabcolsep}{0.25pt}
  \settowidth{\metriclabelwidth}{\mbox{Kimi-VL-A3B-Instruct~\cite{kimiteam2025kimivltechnicalreport}}}
  \begin{tikzpicture}
    \node[inner sep=0pt,outer sep=0pt] (metricgrid) {%
      \begin{minipage}{\linewidth}
  \begin{tabularx}{\linewidth}{@{}l*{14}{>{\centering\arraybackslash}X}@{}}
    \toprule
    \rowcolor{PaperHeader}
    & \multicolumn{5}{c}{Appearance} & \multicolumn{3}{c}{Layout} & \multicolumn{6}{c}{Text} \\
    \cmidrule(lr){2-6}\cmidrule(lr){7-9}\cmidrule(l){10-15}
    \rowcolor{PaperHeader}
    Configuration & \shortstack{SSIM\\$\uparrow$} & \shortstack{LPIPS\\$\downarrow$} & \shortstack{FLIP\\$\downarrow$} & \shortstack{CLIP\\$\uparrow$} & \shortstack{FID\\$\downarrow$} & \shortstack{L-score\\$\uparrow$} & \shortstack{R-F1\\$\uparrow$} & \shortstack{B-IoU\\$\uparrow$} & \shortstack{T-score\\$\uparrow$} & \shortstack{T-F1\\$\uparrow$} & \shortstack{T-IoU\\$\uparrow$} & \shortstack{CER$_m$\\$\downarrow$} & \shortstack{Exact\\$\uparrow$} & \shortstack{CER$_p$\\$\downarrow$} \\
    \midrule
    \multicolumn{15}{@{}l}{\textit{API-accessed}} \\
    \mbox{GPT-5.6 Sol~\cite{openai2026gpt56sol}} & \textcolor{PaperBest}{$\mathbf{0.83}$} & \textcolor{PaperBest}{$\mathbf{0.21}$} & \textcolor{PaperBest}{$\mathbf{0.14}$} & \textcolor{PaperBest}{$\mathbf{0.92}$} & \textcolor{PaperBest}{$\mathbf{9.28}$} & \textcolor{PaperBest}{$\mathbf{0.61}$} & \textcolor{PaperBest}{$\mathbf{0.73}$} & \textcolor{PaperBest}{$\mathbf{0.76}$} & \textcolor{PaperBest}{$\mathbf{0.47}$} & \textcolor{PaperBest}{$\mathbf{0.50}$} & \textcolor{PaperBest}{$\mathbf{0.72}$} & \textcolor{PaperBest}{$\mathbf{0.10}$} & \textcolor{PaperBest}{$\mathbf{0.78}$} & \textcolor{PaperBest}{$\mathbf{0.13}$} \\
    \mbox{Qwen3.7-Plus~\cite{qwen2026qwen37plus}} & \textcolor{PaperSecond}{$0.82$} & \textcolor{PaperSecond}{$0.26$} & \textcolor{PaperSecond}{$0.16$} & \textcolor{PaperSecond}{$0.89$} & \textcolor{PaperSecond}{$13.33$} & \textcolor{PaperSecond}{$0.52$} & \textcolor{PaperSecond}{$0.59$} & \textcolor{PaperSecond}{$0.67$} & \textcolor{PaperSecond}{$0.37$} & \textcolor{PaperSecond}{$0.35$} & $0.66$ & \textcolor{PaperSecond}{$0.16$} & \textcolor{PaperSecond}{$0.72$} & \textcolor{PaperSecond}{$0.16$} \\
    \mbox{MiniMax-M3~\cite{lai2026minimaxsparseattention}} & $0.79$ & $0.36$ & $0.22$ & $0.86$ & $17.69$ & $0.38$ & $0.45$ & $0.52$ & $0.23$ & $0.19$ & $0.66$ & $0.29$ & $0.49$ & $0.23$ \\
    \mbox{Grok-4.5~\cite{xai2026grok45}} & $0.79$ & $0.43$ & $0.24$ & $0.71$ & $62.73$ & $0.14$ & $0.15$ & $0.35$ & $0.10$ & $0.10$ & \textcolor{PaperSecond}{$0.70$} & $0.39$ & $0.41$ & $0.62$ \\
    \midrule
    \multicolumn{15}{@{}l}{\textit{Locally deployed}} \\
    \mbox{Kimi-VL-A3B-Instruct~\cite{kimiteam2025kimivltechnicalreport}} & $0.66$ & $0.56$ & $0.32$ & $0.67$ & $75.66$ & $0.11$ & $0.11$ & $0.34$ & $0.10$ & $0.09$ & \textcolor{PaperBest}{$\mathbf{0.72}$} & $0.53$ & $0.40$ & $0.95$ \\
    \mbox{Qwen3.8-27B~\cite{qwen2026qwen3827b}} & \textcolor{PaperBest}{$\mathbf{0.78}$} & \textcolor{PaperBest}{$\mathbf{0.24}$} & \textcolor{PaperBest}{$\mathbf{0.17}$} & \textcolor{PaperBest}{$\mathbf{0.89}$} & \textcolor{PaperBest}{$\mathbf{11.88}$} & \textcolor{PaperBest}{$\mathbf{0.56}$} & \textcolor{PaperBest}{$\mathbf{0.64}$} & \textcolor{PaperBest}{$\mathbf{0.71}$} & \textcolor{PaperBest}{$\mathbf{0.42}$} & \textcolor{PaperBest}{$\mathbf{0.44}$} & \textcolor{PaperSecond}{$0.69$} & \textcolor{PaperBest}{$\mathbf{0.14}$} & \textcolor{PaperBest}{$\mathbf{0.71}$} & \textcolor{PaperBest}{$\mathbf{0.16}$} \\
    \mbox{Muse-Glimmer-30B~\cite{abtahizadeh2026museglimmer}} & \textcolor{PaperSecond}{$0.76$} & $0.37$ & $0.21$ & \textcolor{PaperSecond}{$0.87$} & $20.52$ & $0.40$ & $0.47$ & $0.55$ & $0.23$ & $0.19$ & $0.66$ & $0.27$ & $0.59$ & $0.25$ \\
    \mbox{Gemma-4-31B-it~\cite{gemmateam2026gemma4}} & \textcolor{PaperSecond}{$0.76$} & \textcolor{PaperSecond}{$0.31$} & \textcolor{PaperSecond}{$0.20$} & \textcolor{PaperSecond}{$0.87$} & \textcolor{PaperSecond}{$17.45$} & \textcolor{PaperSecond}{$0.46$} & \textcolor{PaperSecond}{$0.54$} & \textcolor{PaperSecond}{$0.63$} & \textcolor{PaperSecond}{$0.33$} & \textcolor{PaperSecond}{$0.31$} & $0.66$ & \textcolor{PaperSecond}{$0.18$} & \textcolor{PaperSecond}{$0.62$} & \textcolor{PaperSecond}{$0.23$} \\
    \mbox{InternVL3.5-38B~\cite{wang2025internvl35}} & \textcolor{PaperSecond}{$0.76$} & $0.41$ & $0.23$ & $0.74$ & $45.58$ & $0.26$ & $0.31$ & $0.52$ & $0.18$ & $0.15$ & $0.67$ & $0.35$ & $0.49$ & $0.51$ \\
    \bottomrule
  \end{tabularx}
      \end{minipage}};
    \setlength{\metricgroupoffset}{\dimexpr\metriclabelwidth+\tabcolsep+(\linewidth-\metriclabelwidth)*5/14\relax}
    \draw[PaperDivider,dash pattern=on 2pt off 2pt,line width=.3pt] ([xshift=\metricgroupoffset,yshift=-2pt]metricgrid.north west) -- ([xshift=\metricgroupoffset,yshift=2pt]metricgrid.south west);
    \setlength{\metricgroupoffset}{\dimexpr\metriclabelwidth+\tabcolsep+(\linewidth-\metriclabelwidth)*8/14\relax}
    \draw[PaperDivider,dash pattern=on 2pt off 2pt,line width=.3pt] ([xshift=\metricgroupoffset,yshift=-2pt]metricgrid.north west) -- ([xshift=\metricgroupoffset,yshift=2pt]metricgrid.south west);
  \end{tikzpicture}
\end{table*}

\subsection{Correction on Scientific Figures}

Correcting the native state improves text recovery beyond BBox-JSONL initialization. Across five API configurations on the same $80$ figures, mean text-box F1 rises from $0.27$ to $0.44$ and page CER falls from $0.51$ to $0.43$ (Table~\ref{tab:refinement_details}(b)). Paired resampling supports a positive mean change for these fixed configurations. Supplementary Tables~S8--S9 give all model-specific stages. The gain thus measures correction of an existing native reconstruction, not the difference between code and native representations.

\begin{figure*}[t]
  \centering
  \includegraphics[width=\textwidth]{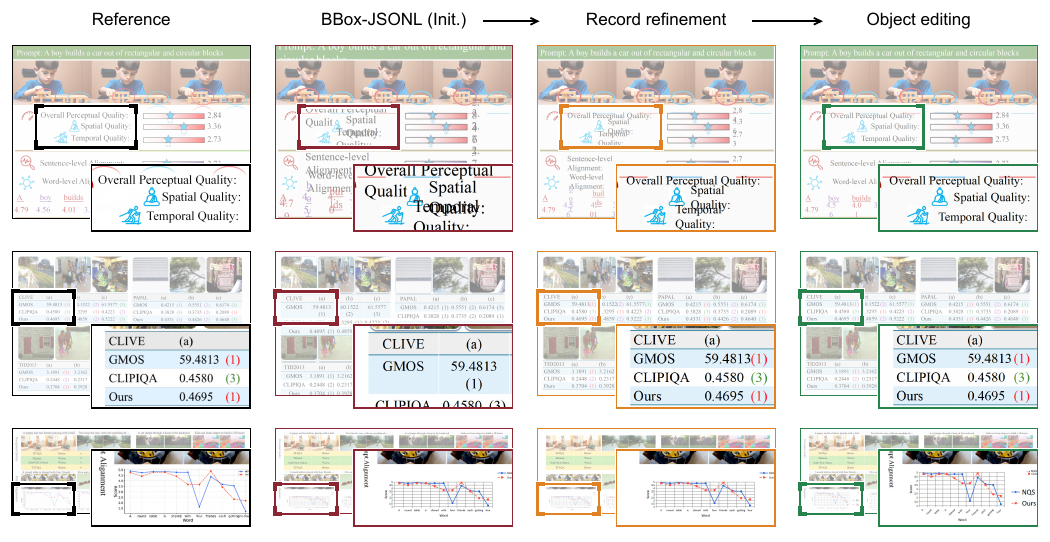}
  \caption{Whole-slide bootstrap correction with Qwen3.8-Max (top) and GPT-5.6 Sol (middle and bottom). Columns show the reference, BBox-JSONL initialization, record refinement, and object editing. The last column precedes iterative correction. Matching frames identify the enlarged regions.}
  \label{fig:qualitative_results}
\end{figure*}

The mean gain hides substantial differences in correction reliability. GPT-5.6 Sol improves text-box F1 on all $80$ figures. MiniMax-M3 improves it on $35$ and worsens it on $34$, with a small increase in mean page CER. Fig.~\ref{fig:qualitative_results} illustrates what changes during bootstrap: record refinement reduces overlapping labels with Qwen3.8-Max and improves table-row placement with GPT-5.6 Sol. 
The curve example retains geometric errors, and 
Supplementary Fig.~S1 shows missing image content with MiniMax-M3. Rendered feedback can guide useful changes without reliably resolving every reconstruction error.

\subsection{Comparison with Reconstruction Tools}

Appearance and text recovery favor different tools (Table~\ref{tab:tool_details}). On $80$ scientific figures, WPS AIPPT~\cite{wps_aippt} has higher SSIM and lower FLIP than Slide2PPT, while Slide2PPT has stronger detected-layout and OCR agreement. For example, text-box F1 is $0.88$ for Slide2PPT and $0.82$ for WPS. We also compare \textsc{scientific-illustrator}~\cite{scientific_illustrator} and \textsc{image-to-editable-ppt-skill}~\cite{image2editableppt_skill}. Slide2PPT uses GPT-5.6 Sol. External versions and budgets are unverified, preventing an isolated comparison of the reconstruction algorithms.

\begin{figure*}[t]
  \centering
  \includegraphics[width=\textwidth]{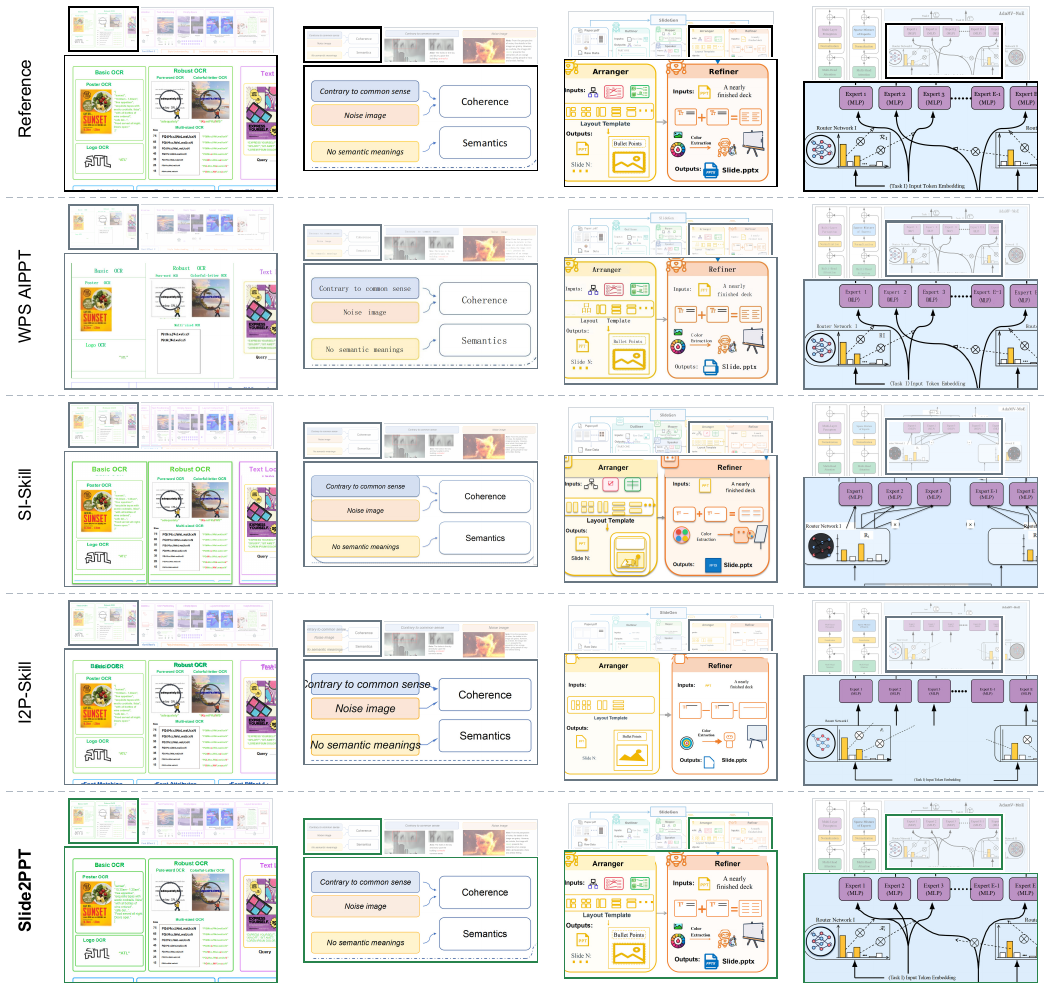}
  \caption{Selected cross-tool reconstructions on inputs distinct from Fig.~\ref{fig:qualitative_results}. Rows show the reference and four tool outputs; columns show four inputs. Slide2PPT uses GPT-5.6 Sol and its retained iterative state; external model versions are unverified. Insets enlarge the same normalized region within each column, preserving each output's orientation and aspect ratio.}
  \label{fig:tool_qualitative}
\end{figure*}

Fig.~\ref{fig:tool_qualitative} shows selected examples with dense panels and structured diagrams. Matched crops expose changes in panel placement, module boundaries, and routing connections. These examples use Slide2PPT's retained iterative outputs on inputs distinct from Fig.~\ref{fig:qualitative_results}. The network example also retains connection errors in Slide2PPT.

\begin{table*}[t]
  \centering
  \caption{Reconstruction tools on $80$ scientific figures; all $14$ quality metrics. Bold marks column optima. SI-Skill and I2P-Skill denote scientific-illustrator and image-to-editable-ppt-skill, respectively. 
  }
    \vspace{-10pt}
  \label{tab:tool_details}
  \small
  \setlength{\tabcolsep}{1pt}
  \settowidth{\metriclabelwidth}{\mbox{WPS AIPPT~\cite{wps_aippt}}}
  \begin{tikzpicture}
    \node[inner sep=0pt,outer sep=0pt] (metricgrid) {%
      \begin{minipage}{\linewidth}
  \begin{tabularx}{\linewidth}{@{}l*{14}{>{\centering\arraybackslash}X}@{}}
    \toprule
    \rowcolor{PaperHeader}
    & \multicolumn{5}{c}{Appearance} & \multicolumn{3}{c}{Layout} & \multicolumn{6}{c}{Text} \\
    \cmidrule(lr){2-6}\cmidrule(lr){7-9}\cmidrule(l){10-15}
    \rowcolor{PaperHeader}
    Tool & \shortstack{SSIM\\$\uparrow$} & \shortstack{LPIPS\\$\downarrow$} & \shortstack{FLIP\\$\downarrow$} & \shortstack{CLIP\\$\uparrow$} & \shortstack{FID\\$\downarrow$} & \shortstack{L-score\\$\uparrow$} & \shortstack{R-F1\\$\uparrow$} & \shortstack{B-IoU\\$\uparrow$} & \shortstack{T-score\\$\uparrow$} & \shortstack{T-F1\\$\uparrow$} & \shortstack{T-IoU\\$\uparrow$} & \shortstack{CER$_m$\\$\downarrow$} & \shortstack{Exact\\$\uparrow$} & \shortstack{CER$_p$\\$\downarrow$} \\
    \midrule
    \mbox{WPS AIPPT~\cite{wps_aippt}} & \textcolor{PaperBest}{$\mathbf{0.85}$} & \textcolor{PaperBest}{$\mathbf{0.15}$} & \textcolor{PaperBest}{$\mathbf{0.13}$} & \textcolor{PaperSecond}{$0.96$} & \textcolor{PaperSecond}{$14.92$} & \textcolor{PaperSecond}{$0.64$} & \textcolor{PaperSecond}{$0.75$} & \textcolor{PaperSecond}{$0.90$} & \textcolor{PaperSecond}{$0.65$} & \textcolor{PaperSecond}{$0.82$} & \textcolor{PaperSecond}{$0.79$} & \textcolor{PaperSecond}{$0.04$} & \textcolor{PaperSecond}{$0.83$} & \textcolor{PaperSecond}{$0.24$} \\
    \mbox{SI-Skill~\cite{scientific_illustrator}} & $0.70$ & $0.37$ & $0.25$ & $0.94$ & $33.93$ & $0.53$ & $0.63$ & $0.75$ & $0.34$ & $0.34$ & $0.67$ & $0.22$ & $0.75$ & $0.38$ \\
    \mbox{I2P-Skill~\cite{image2editableppt_skill}} & $0.72$ & $0.36$ & $0.24$ & $0.92$ & $39.39$ & $0.57$ & $0.65$ & $0.77$ & $0.35$ & $0.37$ & $0.66$ & $0.15$ & $0.78$ & $0.41$ \\
    
    GPT-5.6(Code~\cite{openai2026gpt56sol})
    & $0.61$ & $0.36$ & $0.34$ & $0.80$ & $29.67$
    & $0.56$ & $0.65$ & $0.72$
    & $0.48$ & $0.59$ & $0.61$ & $0.23$ & $0.69$ & $0.40$ \\
    \midrule
    \rowcolor{PaperFocus}
    Slide2PPT(GPT-5.6) & \textcolor{PaperSecond}{$0.80$} & \textcolor{PaperBest}{$\mathbf{0.15}$} & \textcolor{PaperSecond}{$0.15$} & \textcolor{PaperBest}{$\mathbf{0.98}$} & \textcolor{PaperBest}{$\mathbf{13.09}$} & \textcolor{PaperBest}{$\mathbf{0.74}$} & \textcolor{PaperBest}{$\mathbf{0.84}$} & \textcolor{PaperBest}{$\mathbf{0.92}$} & \textcolor{PaperBest}{$\mathbf{0.72}$} & \textcolor{PaperBest}{$\mathbf{0.88}$} & \textcolor{PaperBest}{$\mathbf{0.80}$} & \textcolor{PaperBest}{$\mathbf{0.03}$} & \textcolor{PaperBest}{$\mathbf{0.89}$} & \textcolor{PaperBest}{$\mathbf{0.21}$} \\
    \bottomrule
  \end{tabularx}
      \end{minipage}};
    \setlength{\metricgroupoffset}{\dimexpr\metriclabelwidth+\tabcolsep+(\linewidth-\metriclabelwidth)*5/14+7pt\relax}
    \draw[PaperDivider,dash pattern=on 2pt off 2pt,line width=.3pt] ([xshift=\metricgroupoffset,yshift=-2pt]metricgrid.north west) -- ([xshift=\metricgroupoffset,yshift=2pt]metricgrid.south west);
    \setlength{\metricgroupoffset}{\dimexpr\metriclabelwidth+\tabcolsep+(\linewidth-\metriclabelwidth)*8/14+5pt\relax}
    \draw[PaperDivider,dash pattern=on 2pt off 2pt,line width=.3pt] ([xshift=\metricgroupoffset,yshift=-2pt]metricgrid.north west) -- ([xshift=\metricgroupoffset,yshift=2pt]metricgrid.south west);
  \end{tikzpicture}
\end{table*}

\subsection{Initialization and Correction Analysis}

\begin{table*}[t]
  \centering
  \caption{Initialization strategies and cumulative correction. Entries are mean $\pm$ sample SD across model-level scores. Blocks use different image sets and model cohorts; colors rank scores within each block. Code includes missing-render defaults. Later correction states are cumulative and use additional inference.}
  \vspace{-10pt}
  \label{tab:representation_details}
  \label{tab:refinement_details}
  \small
  \setlength{\tabcolsep}{2.5pt}
  \settowidth{\metriclabelwidth}{+Iterative correction}
  \begin{tikzpicture}
    \node[inner sep=0pt,outer sep=0pt] (metricgrid) {%
      \begin{minipage}{\linewidth}
        \begin{tabularx}{\linewidth}{@{}l*{7}{>{\centering\arraybackslash}X}@{}}
          \toprule
          \rowcolor{PaperHeader}
          & \multicolumn{2}{c}{Appearance}
          & \multicolumn{1}{c}{File production}
          & \multicolumn{2}{c}{Layout}
          & \multicolumn{2}{c}{Text} \\
          \cmidrule(lr){2-3}\cmidrule(lr){4-4}\cmidrule(lr){5-6}\cmidrule(l){7-8}
          \rowcolor{PaperHeader}
          State
          & SSIM $\uparrow$
          & LPIPS $\downarrow$
          & PPTX rate $\uparrow$
          & L-score $\uparrow$
          & R-F1 $\uparrow$
          & T-F1 $\uparrow$
          & CER$_p$ $\downarrow$ \\
          \midrule
          \multicolumn{8}{@{}l}{\textit{(a) Initialization: $108$ random inputs, $10$ models}} \\
          Code
          & $0.36{\pm}0.30$
          & $0.68{\pm}0.29$
          & $0.47{\pm}0.36$
          & $0.21{\pm}0.25$
          & $0.25{\pm}0.29$
          & $0.17{\pm}0.26$
          & $0.62{\pm}0.32$ \\
          JSONL
          & $\textcolor{PaperBest}{\mathbf{0.78}}{\pm}0.04$
          & $\textcolor{PaperSecond}{0.35}{\pm}0.10$
          & $\textcolor{PaperBest}{\mathbf{1.00}}{\pm}0.00$
          & $\textcolor{PaperSecond}{0.37}{\pm}0.16$
          & $\textcolor{PaperSecond}{0.42}{\pm}0.19$
          & $\textcolor{PaperBest}{\mathbf{0.28}}{\pm}0.13$
          & $\textcolor{PaperSecond}{0.38}{\pm}0.35$ \\
          \rowcolor{PaperFocus}
          BBox-JSONL
          & $\textcolor{PaperBest}{\mathbf{0.78}}{\pm}0.05$
          & $\textcolor{PaperBest}{\mathbf{0.34}}{\pm}0.11$
          & $\textcolor{PaperBest}{\mathbf{1.00}}{\pm}0.00$
          & $\textcolor{PaperBest}{\mathbf{0.38}}{\pm}0.17$
          & $\textcolor{PaperBest}{\mathbf{0.44}}{\pm}0.21$
          & $\textcolor{PaperBest}{\mathbf{0.28}}{\pm}0.15$
          & $\textcolor{PaperBest}{\mathbf{0.35}}{\pm}0.28$ \\
          \midrule
          \multicolumn{8}{@{}l}{\textit{(b) Cumulative correction: $80$ scientific figures, $5$ API models}} \\
          Code
          & $0.42{\pm}0.14$
          & $0.65{\pm}0.20$
          & $0.62{\pm}0.15$
          & $0.30{\pm}0.17$
          & $0.35{\pm}0.19$
          & $0.19{\pm}0.25$
          & $0.64{\pm}0.16$ \\
          \rowcolor{PaperFocus}
          BBox-JSONL
          & $0.71{\pm}0.02$
          & $0.44{\pm}0.14$
          & $\textcolor{PaperBest}{\mathbf{1.00}}{\pm}0.00$
          & $0.44{\pm}0.13$
          & $0.53{\pm}0.13$
          & $0.27{\pm}0.16$
          & $0.51{\pm}0.15$ \\
          +Record refinement
          & $\textcolor{PaperSecond}{0.72}{\pm}0.03$
          & $0.41{\pm}0.17$
          & $\textcolor{PaperBest}{\mathbf{1.00}}{\pm}0.00$
          & $0.47{\pm}0.16$
          & $0.56{\pm}0.17$
          & $0.35{\pm}0.26$
          & $0.48{\pm}0.19$ \\
          +Object editing
          & $\textcolor{PaperSecond}{0.72}{\pm}0.04$
          & $\textcolor{PaperSecond}{0.40}{\pm}0.19$
          & $\textcolor{PaperBest}{\mathbf{1.00}}{\pm}0.00$
          & $\textcolor{PaperSecond}{0.48}{\pm}0.17$
          & $\textcolor{PaperSecond}{0.57}{\pm}0.18$
          & $\textcolor{PaperSecond}{0.38}{\pm}0.31$
          & $\textcolor{PaperSecond}{0.45}{\pm}0.20$ \\
          \rowcolor{PaperFocus}
          +Iterative correction
          & $\textcolor{PaperBest}{\mathbf{0.73}}{\pm}0.05$
          & $\textcolor{PaperBest}{\mathbf{0.37}}{\pm}0.19$
          & $\textcolor{PaperBest}{\mathbf{1.00}}{\pm}0.00$
          & $\textcolor{PaperBest}{\mathbf{0.49}}{\pm}0.19$
          & $\textcolor{PaperBest}{\mathbf{0.58}}{\pm}0.20$
          & $\textcolor{PaperBest}{\mathbf{0.44}}{\pm}0.34$
          & $\textcolor{PaperBest}{\mathbf{0.43}}{\pm}0.22$ \\
          \bottomrule
        \end{tabularx}
      \end{minipage}};

    \setlength{\metricgroupoffset}{\dimexpr\metriclabelwidth+\tabcolsep+(\linewidth-\metriclabelwidth)*2/7\relax}
    \draw[PaperDivider,dash pattern=on 2pt off 2pt,line width=.3pt]
      ([xshift=\metricgroupoffset,yshift=-2pt]metricgrid.north west)
      --
      ([xshift=\metricgroupoffset,yshift=2pt]metricgrid.south west);

    \setlength{\metricgroupoffset}{\dimexpr\metriclabelwidth+\tabcolsep+(\linewidth-\metriclabelwidth)*3/7\relax}
    \draw[PaperDivider,dash pattern=on 2pt off 2pt,line width=.3pt]
      ([xshift=\metricgroupoffset,yshift=-2pt]metricgrid.north west)
      --
      ([xshift=\metricgroupoffset,yshift=2pt]metricgrid.south west);

    \setlength{\metricgroupoffset}{\dimexpr\metriclabelwidth+\tabcolsep+(\linewidth-\metriclabelwidth)*5/7\relax}
    \draw[PaperDivider,dash pattern=on 2pt off 2pt,line width=.3pt]
      ([xshift=\metricgroupoffset,yshift=-2pt]metricgrid.north west)
      --
      ([xshift=\metricgroupoffset,yshift=2pt]metricgrid.south west);
  \end{tikzpicture}

  \vspace{-12pt}
\end{table*}

The clearest benefit of typed output is file reliability. On $108$ images across ten configurations, both JSONL settings produce readable PPTX files in all $1{,}080$ model--image cases, while Code produces $506$ (rate $0.47$; Table~\ref{tab:representation_details}(a)). This reliability gain does not imply uniformly better reconstruction: GPT-5.6 Sol with Code reaches SSIM $0.84$ and layout score $0.77$, versus $0.83$ and $0.60$ for BBox-JSONL. Sharing the rendering implementation removes one source of failure while leaving content and spatial prediction to the model.

Bounding-box guidance provides a useful but imperfect spatial prior. Relative to direct JSONL, it raises mean layout score from $0.37$ to $0.38$ and region F1 from $0.42$ to $0.44$. Layout score improves for Qwen3.7-Plus but declines for Kimi-VL-A3B-Instruct. The mixed effect supports allowing record prediction to revise the detected boxes instead of fixing the initial decomposition.

Gains accrue across the bootstrap stages: mean text-box F1 rises from $0.27$ to $0.35$ after record refinement and to $0.38$ after object editing, reaching $0.44$ after iteration. Each state uses additional inference, so this comparison measures accumulated correction rather than the benefit of a single mechanism at equal cost. The initialization and correction blocks also use different inputs and model cohorts, and their means cannot be compared as paired results.

\subsection{Native Editing Interfaces}
\label{sec:objects}

The native representation exposes editing interfaces beyond geometric primitives. Across five API configurations on $80$ figures, structured outputs contain $77$ native charts and $126$ native tables, compared with $24$ and $40$ for Code. In the GPT-5.6 Sol tool comparison, Slide2PPT produces both charts and tables, while WPS produces more tables but no charts. Supplementary Table~S16 reports all object types. These counts establish availability, not correct recovery: over-segmentation or type errors can increase the inventory, and the records do not identify gains attributable to iteration. Series, cells, and relationships still require reference matching and executed-edit tests.

%% file: sec/07_discussion.tex
\section{Discussion and Limitations}
\label{sec:discussion}

A shared native state makes visual errors addressable through document properties. Its value depends, however, on what those properties represent. A chart with editable series can still encode the wrong values, and a native table can contain the wrong cell structure. Our measurements establish file production, rendered agreement, and object availability. Source-matched evaluation and actual editing tasks are needed to test whether those interfaces preserve the intended content and relationships.

Regional selection limits which records can change, but excludes boundary-crossing objects and does not guarantee unchanged pixels outside the window. Whole-slide correction can revisit these dependencies, at the cost of allowing broader changes. The current correction study covers scientific figures, uses more inference at later stages, and selects candidates with visual metrics also used in evaluation. It therefore leaves open the gains under matched budgets, independent quality assessment, and correction of ordinary slides.

%% file: sec/06_conclusion.tex
\section{Conclusion}

Slide2PPT makes native PowerPoint objects the common target of image reconstruction and visual correction. A fixed renderer exposes consistent editing properties, while discrepancy-guided updates revise their content and placement. Experiments show that typed prediction reliably produces readable files and that correction improves average reconstruction quality on scientific figures. Recovering the right native structure and data remains a separate requirement for successful downstream editing.

%% file: main.bib
@ARTICLE{1284395,
  author={Wang, Zhou and Bovik, Alan C. and Sheikh, Hamid R. and Simoncelli, Eero P.},
  journal={IEEE Trans. Image Process.}, 
  title={Image quality assessment: from error visibility to structural similarity}, 
  year={2004},
  volume={13},
  number={4},
  pages={600--612},
  
  doi={10.1109/TIP.2003.819861},
  url={https://ece.uwaterloo.ca/~z70wang/publications/ssim.html}
}

@article{andersson2020flip,
  title={{FLIP}: A Difference Evaluator for Alternating Images},
  author={Andersson, Pontus and Nilsson, Jim and Akenine-M{\"o}ller, Tomas and Oskarsson, Magnus and {\AA}str{\"o}m, Kalle and Fairchild, Mark D},
  journal={Proc. ACM Comput. Graph. Interact. Tech.},
  volume={3},
  number={2},
  pages={15:1--15:23},
  year={2020},
  doi={10.1145/3406183}
}

@article{cui2025paddleocr30technicalreport,
      title={{PaddleOCR} 3.0 Technical Report}, 
      author={Cheng Cui and Ting Sun and Manhui Lin and Tingquan Gao and Yubo Zhang and Jiaxuan Liu and Xueqing Wang and Zelun Zhang and Changda Zhou and Hongen Liu and Yue Zhang and Wenyu Lv and Kui Huang and Yichao Zhang and Jing Zhang and Jun Zhang and Yi Liu and Dianhai Yu and Yanjun Ma},
      year={2025},
      eprint={2507.05595},
      archivePrefix={arXiv},
      primaryClass={cs.CV},
      url={https://arxiv.org/abs/2507.05595},
  journal={arXiv:2507.05595}
}

@article{fan2026refigbench,
  title = {{ReFigBench}: Benchmarking Scientific Figure Reconstruction as Editable {PowerPoint} Artifacts},
  author = {Fan, Liyang and Wei, Chi and Li, Yitai and Bi, Xinping and Chen, Guhong and Sun, Chenghao and Yang, Haoxiang and Li, Qingwen and Yan, Kai and Li, Hong and Li, Bo},
  journal = {arXiv:2609.18844},
  year = {2026},
  eprint = {2609.18844},
  archivePrefix = {arXiv},
  url = {https://arxiv.org/abs/2609.18844}
}

@inproceedings{fu2022doc2ppt,
  title={{DOC2PPT}: Automatic Presentation Slides Generation from Scientific Documents},
  author={Fu, Tsu-Jui and Wang, William Yang and McDuff, Daniel and Song, Yale},
  booktitle={AAAI},
  volume={36},
  number={1},
  pages={634--642},
  year={2022},
  doi={10.1609/aaai.v36i1.19943},
  url={https://ojs.aaai.org/index.php/AAAI/article/view/19943}
}

@inproceedings{ge2025autopresent,
  title={{AutoPresent}: Designing Structured Visuals from Scratch},
  author={Ge, Jiaxin and Wang, Zora Zhiruo and Zhou, Xuhui and Peng, Yi-Hao and Subramanian, Sanjay and Tan, Qinyue and Sap, Maarten and Suhr, Alane and Fried, Daniel and Neubig, Graham and Darrell, Trevor},
  booktitle={CVPR},
  pages={2902--2911},
  year={2025},
  url={https://openaccess.thecvf.com/content/CVPR2025/html/Ge_AutoPresent_Designing_Structured_Visuals_from_Scratch_CVPR_2025_paper.html}
}

@article{gonzalez2026images2slides,
  title = {From Dead Pixels to Editable Slides: Infographic Reconstruction into Native {Google Slides} via Vision-Language Region Understanding},
  author = {Gonzalez, Leonardo},
  journal = {arXiv:2602.07645},
  year = {2026},
  eprint = {2602.07645},
  archivePrefix = {arXiv},
  url = {https://arxiv.org/abs/2602.07645}
}

@article{hazimeh2025slider,
  title = {Semantic Document Derendering: {SVG} Reconstruction via Vision-Language Modeling},
  author = {Hazimeh, Adam and Wang, Ke and Collier, Mark and Baechler, Gilles and Kokiopoulou, Efi and Frossard, Pascal},
  journal = {arXiv:2511.13478},
  year = {2025},
  eprint = {2511.13478},
  archivePrefix = {arXiv},
  url = {https://arxiv.org/abs/2511.13478}
}

@inproceedings{heusel2017gans,
  title={{GANs} Trained by a Two Time-Scale Update Rule Converge to a Local {Nash} Equilibrium},
  author={Heusel, Martin and Ramsauer, Hubert and Unterthiner, Thomas and Nessler, Bernhard and Hochreiter, Sepp},
  
  volume={30},
  year={2017},
  booktitle={NeurIPS},
  url={https://proceedings.neurips.cc/paper/2017/hash/8a1d694707eb0fefe65871369074926d-Abstract.html}
}

@article{huang2025pptbench,
  title={{PPTBench}: Towards Holistic Evaluation of Large Language Models for {PowerPoint} Layout and Design Understanding},
  author={Huang, Zheng and Liu, Xukai and Hu, Tianyu and Zhang, Kai and Liu, Ye},
  journal={arXiv:2512.02624},
  year={2025},
  url={https://arxiv.org/abs/2512.02624}
}

@misc{image2editableppt_skill,
  author = {{ningzimu}},
  title  = {Image to Editable {PPT} Skill},
  howpublished = {GitHub repository},
  year   = {2026},
  url    = {https://github.com/ningzimu/image-to-editable-ppt-skill},
  note   = {Accessed 2026-11-16}
}

@inproceedings{radford2021learning,
  title={Learning transferable visual models from natural language supervision},
  author={Radford, Alec and Kim, Jong Wook and Hallacy, Chris and Ramesh, Aditya and Goh, Gabriel and Agarwal, Sandhini and Sastry, Girish and Askell, Amanda and Mishkin, Pamela and Clark, Jack and Krueger, Gretchen and Sutskever, Ilya},
  booktitle={ICML},
  pages={8748--8763},
  year={2021},
  url={https://proceedings.mlr.press/v139/radford21a.html}
}

@misc{scientific_illustrator,
  author = {{icebird1998}},
  title  = {Scientific Illustrator},
  howpublished = {GitHub repository},
  year   = {2026},
  url    = {https://github.com/icebird1998/scientific-illustrator},
  note   = {Accessed 2026-11-16}
}

@inproceedings{shi2024chartmimic,
  title = {{ChartMimic}: Evaluating {LMM}'s Cross-Modal Reasoning Capability via Chart-to-Code Generation},
  author = {Yang, Cheng and Shi, Chufan and Liu, Yaxin and Shui, Bo and Wang, Junjie and Jing, Mohan and Xu, Linran and Zhu, Xinyu and Li, Siheng and Zhang, Yuxiang and Liu, Gongye and Nie, Xiaomei and Cai, Deng and Yang, Yujiu},
  
  year = {2025},
  eprint = {2406.09961},
  archivePrefix = {arXiv},
  url = {https://proceedings.iclr.cc/paper_files/paper/2025/hash/42806406dd99e30c3796bc98b2670fa2-Abstract-Conference.html},
  booktitle={ICLR}
}

@inproceedings{si2024design2code,
  title = {{Design2Code}: Benchmarking Multimodal Code Generation for Automated Front-End Engineering},
  author = {Si, Chenglei and Zhang, Yanzhe and Li, Ryan and Yang, Zhengyuan and Liu, Ruibo and Yang, Diyi},
  
  year = {2025},
  eprint = {2403.03163},
  archivePrefix = {arXiv},
  url = {https://aclanthology.org/2025.naacl-long.199/},
  booktitle={NAACL},
  pages={3956--3974},
  doi={10.18653/v1/2025.naacl-long.199}
}

@inproceedings{tang2025slidecoder,
  title={{SlideCoder}: Layout-aware {RAG}-enhanced Hierarchical Slide Generation from Design},
  author={Tang, Wenxin and Xiao, Jingyu and Jiang, Wenxuan and Xiao, Xi and Wang, Yuhang and Tang, Xuxin and Li, Qing and Ma, Yuehe and Liu, Junliang and Tang, Shisong and Lyu, Michael R.},
  booktitle={EMNLP},
  pages={9015--9039},
  year={2025},
  doi={10.18653/v1/2025.emnlp-main.458},
  url={https://aclanthology.org/2025.emnlp-main.458/}
}

@article{tian2026redeck,
  title = {{ReDeck}: Step-Level Render-Grounded Refinement for Document-to-Slide Generation},
  author = {Tian, Muzhao and Zeng, Zezi and Yang, Yifan and Gao, Xin and Li, Yan and Huang, Zisu and Wang, Xiaohua and Lv, Changze and Cheng, Mingxi and Liu, Bei and Qiu, Kai and Dai, Qi and Chen, Dong and Dong, Yue and Zheng, Xiaoqing and Li, Ji and Luo, Chong},
  journal = {arXiv:2609.00194},
  year = {2026},
  eprint = {2609.00194},
  archivePrefix = {arXiv},
  url = {https://arxiv.org/abs/2609.00194}
}

@article{wang2026pptbenchreconstruction,
  title = {{PPTBench}: Can Coding Agents Reconstruct the Visual World through Structured, Editable Slides},
  author = {Wang, Xiaoqiu and Chi, Yizhe and Li, Wenyi and Hong, Deyao and Shan, Zhihan and Gao, Mingju and Yang, Kaisen and Zheng, Youjie and Xiao, Calvin and Na, Qinhuai},
  journal = {arXiv:2609.29718},
  year = {2026},
  eprint = {2609.29718},
  archivePrefix = {arXiv},
  url = {https://arxiv.org/abs/2609.29718}
}

@misc{wps_aippt,
  author = {{Kingsoft Office}},
  title  = {{WPS AIPPT}: Image-to-{PowerPoint} Conversion},
  howpublished = {Official service},
  year   = {2026},
  url    = {https://aippt.wps.cn/aippt/convert-ppt/home},
  note   = {Accessed 2026-11-16}
}

@inproceedings{zhang2018perceptual,
  title={The Unreasonable Effectiveness of Deep Features as a Perceptual Metric},
  author={Zhang, Richard and Isola, Phillip and Efros, Alexei A and Shechtman, Eli and Wang, Oliver},
  booktitle={CVPR},
  year={2018},
  pages={586--595},
  url={https://openaccess.thecvf.com/content_cvpr_2018/html/Zhang_The_Unreasonable_Effectiveness_CVPR_2018_paper.html}
}

@article{zhao2024doclayout,
  title={{DocLayout-YOLO}: Enhancing Document Layout Analysis through Diverse Synthetic Data and Global-to-Local Adaptive Perception},
  author={Zhao, Zhiyuan and Kang, Hengrui and Wang, Bin and He, Conghui},
  journal={arXiv:2410.12628},
  year={2024},
  url={https://arxiv.org/abs/2410.12628}
}

@inproceedings{zheng2025pptagent,
  title={{PPTAgent}: Generating and Evaluating Presentations Beyond Text-to-Slides},
  author={Zheng, Hao and Guan, Xinyan and Kong, Hao and Zhang, Wenkai and Zheng, Jia and Zhou, Weixiang and Lin, Hongyu and Lu, Yaojie and Han, Xianpei and Sun, Le},
  booktitle={EMNLP},
  pages={14402--14418},
  year={2025},
  doi={10.18653/v1/2025.emnlp-main.728},
  url={https://aclanthology.org/2025.emnlp-main.728/}
}

@inproceedings{zheng2026slideforge,
  title = {{SlideForge}: Controllable Editing of Slides as Structured Artifacts},
  author = {Zheng, Haozhen and Wang, Fulin and Xiong, Tianhu and Yu, Yingjie and Qian, Shengyi and Yu, Hanchao and Schwing, Alex and Nahrstedt, Klara and Wu, Mingyuan},
  
  year = {2026},
  eprint = {2609.03109},
  archivePrefix = {arXiv},
  url = {https://aclanthology.org/2026.emnlp-main.1690/},
  booktitle={EMNLP},
  pages={38429--38471}
}

@misc{openai2026gpt56sol,
  author = {{OpenAI}},
  title = {{GPT-5.6 Sol} Model},
  year = {2026},
  howpublished = {OpenAI API documentation},
  url = {https://developers.openai.com/api/docs/models/gpt-5.6-sol},
  note = {Accessed 2026-11-16}
}

@misc{qwen2026qwen38max,
  author = {{Qwen Team}},
  title = {{Qwen3.8-Max}},
  year = {2026},
  howpublished = {QwenCloud model documentation},
  url = {https://docs.qwencloud.com/models/qwen3-8-max},
  note = {Accessed 2026-11-16}
}

@misc{qwen2026qwen37plus,
  author = {{Qwen Team}},
  title = {{Qwen3.7-Plus}},
  year = {2026},
  howpublished = {QwenCloud model documentation},
  url = {https://docs.qwencloud.com/models/qwen3-7-plus},
  note = {Accessed 2026-11-16}
}

@article{lai2026minimaxsparseattention,
  author = {Xunhao Lai and Weiqi Xu and Yufeng Yang and others},
  title = {{MiniMax} Sparse Attention},
  journal = {arXiv:2606.13392},
  year = {2026},
  url = {https://arxiv.org/abs/2606.13392}
}

@misc{xai2026grok45,
  author = {{xAI}},
  title = {{Grok 4.5}},
  year = {2026},
  howpublished = {API model documentation},
  url = {https://docs.x.ai/developers/models/grok-4.5},
  note = {Accessed 2026-11-16}
}

@article{kimiteam2025kimivltechnicalreport,
  author = {{Kimi Team} and Angang Du and Bohong Yin and others},
  title = {{Kimi-VL} Technical Report},
  journal = {arXiv:2504.07491},
  year = {2025},
  url = {https://arxiv.org/abs/2504.07491}
}

@misc{qwen2026qwen3827b,
  author = {{Qwen Team}},
  title = {{Qwen3.8-27B} Model Card},
  year = {2026},
  howpublished = {Hugging Face},
  url = {https://huggingface.co/Qwen/Qwen3.8-27B},
  note = {Accessed 2026-11-16}
}

@article{abtahizadeh2026museglimmer,
  author = {Arash Ebrahim Abtahizadeh and Muna Aghamelu and Adil Ahmed and others},
  title = {{Muse Glimmer 30B}: An On-Device Agentic Model},
  journal = {arXiv:2610.24628},
  year = {2026},
  url = {https://arxiv.org/abs/2610.24628}
}

@article{gemmateam2026gemma4,
  author = {{Gemma Team} and El Abd, Sherif and Vaibhav Aggarwal and others},
  title = {{Gemma 4} Technical Report},
  journal = {arXiv:2607.02770},
  year = {2026},
  url = {https://arxiv.org/abs/2607.02770}
}

@article{wang2025internvl35,
  author = {Weiyun Wang and Zhangwei Gao and Lixin Gu and others},
  title = {{InternVL3.5}: Advancing Open-Source Multimodal Models in Versatility, Reasoning, and Efficiency},
  journal = {arXiv:2508.18265},
  year = {2025},
  url = {https://arxiv.org/abs/2508.18265}
}
